%% file: dppp.tex
\documentclass[]{article}
\usepackage{proceed2e}
\usepackage{colonequals}
\usepackage{amsmath, amssymb, amsthm, amsfonts}
\usepackage{mathrsfs}
\usepackage[round,sort&compress]{natbib}
\usepackage{algorithm}
\usepackage{algpseudocode}
\usepackage{lipsum}
\usepackage{bbm}
\usepackage{listings}
\usepackage{natbib}
\usepackage{textcomp}
\usepackage{caption, subcaption}
\usepackage{wrapfig}
\usepackage{graphicx}
\usepackage{color}
\usepackage[hyphens]{url}
\usepackage{eqparbox,array}
\usepackage{tikz}

\DeclareFontShape{OT1}{cmtt}{bx}{n}{<5><6><7><8><9><10><10.95><12><14.4><17.28><20.74><24.88>cmttb10}{}

\newcommand{\Dom}[1]{\mathcal{#1}}
\newcommand{\Conts}{\Dom{C}}
\newcommand{\Choices}{\Dom{R}}
\newcommand{\Calls}{\Dom{S}}
\newcommand{\PartialResults}{\Dom{X}}
\newcommand{\Values}{\Dom{V}}
\newcommand{\ProbMeasures}{\Dom{P(V)}}

\algtext*{EndWhile}
\algtext*{EndIf}
\algtext*{EndFor}

\newcommand{\Val}{\mathcal{V}}
\newtheorem{mydef}{Definition}

\newcommand{\Interpreter}{\mathcal{I}}

\newcommand{\prodnode}{n_{\mathrm{prod}}}
\newcommand{\refnode}{n_{\mathrm{ref}}}
\newcommand{\prevnode}{n_{\mathrm{prev}}}
\newcommand{\prevweight}{w_{\mathrm{prev}}}
\newcommand{\rootnode}{n_{\mathrm{root}}}
\newcommand{\curnode}{n_{\mathrm{cur}}}
\newcommand{\initstate}{x_{\mathrm{init}}}

\title{A Dynamic Programming Algorithm for Inference \\ in Recursive Probabilistic Programs}

\author{ {\bf Andreas Stuhlm\"{u}ller} \\
Department of Brain and Cognitive Sciences \\
Massachusetts Institute of Technology
\And
{\bf Noah D.~Goodman}   \\
Department of Psychology \\
Stanford University
}

\begin{document}

\maketitle

\begin{abstract}

We describe a dynamic programming algorithm for computing the marginal distribution of discrete probabilistic programs. This algorithm takes a functional interpreter for an arbitrary probabilistic programming language and turns it into an efficient marginalizer. 
Because direct caching of sub-distributions is impossible in the presence of recursion,
we build a graph of dependencies between sub-distributions. This \emph{factored sum-product network} makes (potentially cyclic) dependencies between subproblems explicit, and corresponds to a system of equations for the marginal distribution. We solve these equations by fixed-point iteration in topological order. We illustrate this algorithm on examples used in teaching probabilistic models, computational cognitive science research, and game theory.

\end{abstract}

\section{INTRODUCTION}

Probabilistic programming allows rapid prototyping of complexly structured probabilistic models without requiring the design of model-specific inference algorithms. This makes probabilistic programs attractive for scientific research: when hypotheses are formalized as programs, it is possible to quickly explore the space of hypotheses. The same features make probabilistic programs compelling for education: students can focus on understanding modeling and inference {\em patterns} before they need to learn about inference {\em implementations}.

However, the performance of current inference algorithms for generic probabilistic programs can vary greatly between models, even for models with a very small number of random choices. This presents an obstacle to the use of probabilistic programs in research and teaching. In fact, many of the models used in these domains are small enough that exact computation is feasible in principle, but they often exhibit patterns, such as nested conditioning, that make naive enumeration intractable. 

In this paper we develop a generic dynamic programming algorithm, which expands the applicability of exact inference for probabilistic programs.
Given an interpreter for an arbitrary probabilistic programming language and a discrete probabilistic program, this algorithm computes the marginal distribution of the program---i.e., its distribution on return values---while sharing subcomputations where possible. By viewing conditioning as marginalization of a rejection sampler, this  captures the full range of probabilistic operations over arbitrary models.

The key obstacle to dynamic programming, which is neither present in caching deterministic interpreters nor in dynamic programming algorithms for more restricted model classes, is the possibility of stochastic self-recursion: an interpreter call with particular arguments can result in a call with the same arguments. Figure \ref{fig:probprog_recursive}a shows a program that exhibits this property. This is not a corner case: for instance, all models that implement conditioning via rejection sampling have this property (Figure \ref{fig:probprog_recursive}b).

To make dynamic programming possible in the presence of recursion, we first compile the given probabilistic program to an intermediate representation that reifies dependencies between sub-distributions. We then compute the marginal distribution from this representation.
Our intermediate representation is a generalization of sum-product networks \citep[][]{Poon:2011vd} that makes dependencies---including recursive dependencies---explicit: a {\em factored sum-product network} (FSPN). While computing the distribution implied by a sum-product network is linear in the size of the network, FSPNs are more difficult to solve in general. We solve FSPNs by clustering their vertices into strongly connected components and by solving each component using fixed-point iteration.

In the following, we first describe the structures our algorithm operates on: probabilistic programs, their interpreters, and FSPNs. We then present the two steps of our algorithm, compilation of programs to FSPNs and computation of marginal distributions given a FSPN. We demonstrate the algorithm on examples used in teaching, cognitive science research, and game theory, and explain what makes it attractive in each case. We relate the algorithm to the literature and conclude with future research directions.

\section{PROBABILISTIC PROGRAMS}

\begin{figure}[t]
  \begin{subfigure}[b]{0.45\textwidth}
    \begin{lstlisting}
(define (game player)
  (if (flip .6)
      (not (game (not player)))
      (if player
          (flip .2) 
          (flip .7))))

(game true)
\end{lstlisting}
  \label{fig:probprog_game}
  \caption{A simple game}
  \end{subfigure}
  \begin{subfigure}[b]{0.5\textwidth}
   \begin{lstlisting}
(define (rejection joint condition?)
  (let ([sample (joint)])
    (if (condition? sample)
        sample
        (rejection joint condition?))))
       \end{lstlisting}
  \caption{Conditioning}
  \label{fig:probprog_conditioning}
  \end{subfigure}    
\caption{Recursive probabilistic programs}
\label{fig:probprog_recursive}
\vskip -1em
\end{figure}

A probabilistic program is a program in a language with primitives for sampling from distributions such as Bernoulli and multinomial. Probabilistic programs describe generative models and thus denote distributions. An interpreter specifies this denotation by implementing a process that, given a program, generates samples from the program's distribution.
For example, an interpreter for the Church language \citep[][]{Goodman:2008uq} takes a program expression and environment, and returns a sample from the program's distribution on Church values. This sample is generated using recursive calls to the interpreter, with each subcall defining a distribution on values and resulting in a sample from this sub-distribution.

The problem of inference for generative models is commonly formulated in terms of a conditioning. However, for any conditional distribution there is an equivalent unconditioned model that samples outcomes with the same probabilities. Inference can be understood as the problem of marginalization of this new model.

A simple way to construct a generative model which samples from some conditional distribution is via {\it rejection sampling}.  Figure \ref{fig:probprog_recursive}b shows how this works in the Church language.  Assume that the procedure \texttt{joint} draws samples from some joint distribution and that {\tt condition?} is a predicate which checks whether some condition holds for each sample.
The recursive procedure  \texttt{rejection} draws samples from \texttt{joint} conditional on {\tt condition?}. Crucially, this procedure makes use of no special conditioning operator but samples directly from the conditional distribution of interest.

Of course, drawing conditional samples using \texttt{rejection} is very inefficient: In general, we may have to tolerate an exponential number of rejected samples before the condition is satisfied. However, if we could efficiently marginalize the \texttt{rejection} procedure, eliminating all zero-probability paths, then we would have solved our target inference problem.

For many probabilistic programs, efficient marginalization of this sort is possible.
We are interested in programs for which many different executions share substructure. Problems with this character are classically amenable to dynamic programming.   
Recursive programs, like \texttt{rejection}, which involve multiple executions of the same procedure application, provide a particularly rich opportunity to exploit shared substructure.
We will focus on these cases in the examples below.

\section{FACTORED SUM-PRODUCT NETWORKS}
\label{sec:fspn}

\begin{figure*}[t]
  \includegraphics[scale=.8]{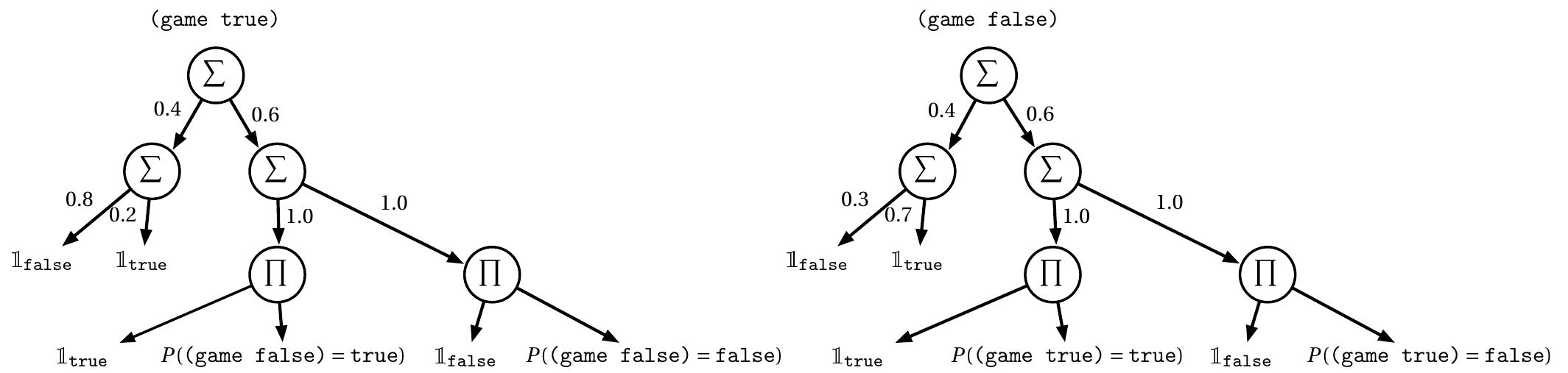}
  \caption{The factored sum-product network corresponding to the \texttt{game} program (Figure \ref{fig:probprog_recursive}a), showing sums, products, indicators, and reference nodes. $P(r=v)$ references the probability of value $v$ under node $r$.}
  \label{fig:rspn-program}
\end{figure*}

In the process of computing the marginal distribution for a given probabilistic program, we use {\em factored sum-product networks} as an intermediate representation between original program and marginal distribution. Like sum-product networks \citep{Poon:2011vd}, this representation factors out all ``deterministic'' computation, leaving only probability calculations, i.e., sums and products. In addition, it makes explicit the dependencies between the distributions resulting from subcomputations.

\begin{mydef}
  A factored sum-product network (FSPN) over variables $x_1,\dots,x_d$ is a directed graph with a uniquely labeled root node $r$.
  The internal nodes are sums and products.
  The leaves are indicators $x_1,\dots,x_d$ and $\bar{x}_1,\dots,\bar{x}_d$, and
  reference nodes $(y, \vec{x})$, where $y$ is another node and $\vec{x}$ a vector of indicator values.
  Each edge $(i, j)$ from a sum node $i$ has a non-negative weight $w_{ij}$.
\end{mydef}

Let $\mathrm{Ch}(y)$ denote the children of node $y$.
The value $\Val(y, \vec{x})$ of a node $y$ is defined as 
$\sum_{z \in \mathrm{Ch}(y)} w_{yz} \Val(z, \vec{x})$ if $y$ is a sum, 
as $\prod_{z \in \mathrm{Ch}(y)} \Val(z, \vec{x})$ if $y$ is a product,
as $\mathbbm{1}_{\vec{x}_j = x_j}$ if $y$ is an indicator $x_j$, and 
as $\Val(z, \vec{w})$ if $y$ is a reference $(z, \vec{w})$.
This defines a system of equations.
We denote the factored sum-product network $F$ as a function of the indicator variables
$\vec{x} = (x_1,\dots,x_d, \bar{x}_1,\dots,\bar{x}_d)$
by $F(\vec{x}) = \Val(r, \vec{x})$. For any given $\vec{x}$, the value of the FSPN is the solution to the system of equations $F(\vec{x})$ (if a unique solution exists).

\section{ALGORITHM}

\subsection{Overview}

Our algorithm solves the following problem: Given a functional interpreter for a probabilistic programming language, how can we build an efficient marginalizer? In other words: How can we turn a universal sampler into a generic dynamic programming algorithm?

If the interpreter were deterministic, we could just memoize it. If it were an interpreter for a stochastic language that does not allow self-recursion, i.e., where an interpreter call can never result in an interpreter call with the same arguments, we could use the interpreter to recursively compute and cache the distribution for each unique interpreter call. For languages that allow self-recursion, direct caching is no longer feasible, since it could lead to infinite regress.

On a high level, our approach is this: our algorithm takes as input the interpreter and a program, and builds a factored sum-product network, which can then be solved using methods such as fixed-point iteration. We intercept any recursive calls the interpreter makes to itself and any calls it makes to its source of randomness, and build network structure that reflects these calls. The FSPN constructed in this way describes the marginal distribution of the interpreted program.

\subsection{Interpreters as Factored Coroutines}
\label{sec:coroutine}

As a prerequisite for the description of our algorithm, we now present mathematical objects that formalize the idea of an interpreter as a coroutine.

Let $\mathcal{V}$ be a countable domain of values that a probabilistic program can return, and let $\mathcal{P(V)}$ be the domain of probability measures on $\mathcal{V}$.

Let $\Conts$, $\Choices$, $\Calls$, and $\PartialResults$ denote the (as yet undefined) domains of {\em continuations}, {\em random choices}, {\em subcalls}, and {\em partial results}. Loosely speaking, (1) continuations are functions from values to partial results, (2) random choices are pairs of continuations and distributions on values, (3) subcalls are pairs of continuations and partial results, and (4) partial results are either values, random choices, or subcalls.

Formally, let $\Conts, \Choices, \Calls,$ and $\PartialResults$ be the smallest domains satisfying the recursive domain equations:
\begin{align}
  \Conts = \Values \to \PartialResults \\
  \Choices = \Conts \times \ProbMeasures \\
  \Calls = \Conts \times \PartialResults \\
  \PartialResults = \Values \cup \Choices \cup \Calls
\end{align}
An interpreter is a function from partial results to partial results.

\subsection{Compiling Programs to FSPNs}

\texttt{BuildFSPN}, shown in Algorithm 1, takes as arguments an interpreter in factored coroutine form and an initial interpreter argument $x_{\mathrm{init}}$. The algorithm steps through all possible execution paths while building the corresponding factored sum-product network, but avoiding duplicate evaluation of subproblems. We first describe three ingredients for this procedure---the task queue, constant-time subcall identification, and factorization grain---then the algorithm \texttt{BuildFSPN}.

{\bf Task queue.} In programs with self-recursive calls, the exploration order of different execution paths can be highly constrained.
For example, in order to evaluate the first \texttt{if}-branch of \texttt{(game true)}
in Figure \ref{fig:probprog_recursive}a, we need to know at least one of the return values of \texttt{(game (not true))}, but these in turn depend on the return values of \texttt{(game true)}.
In order to let program exploration be guided by what return values are known, we maintain a map \texttt{terminals}, which maps each root node to all known terminal values reachable from it, and a map \texttt{callbacks}, which maps each root node to a list of callbacks. A callback is a pair of a node $n$ and a continuation $c$. When a new terminal $v$ is found below a root node associated with callback $(n, c)$, the call $c(v)$ is used to continue evaluation and network building in the original context.

{\bf Constant-time subcall identification.} At each subcall, we need to determine whether the subcall is new or whether it has already been assigned a FSPN node. For the algorithm to have constant-time overhead over steps of the underlying interpreter, it is crucial that this computation takes place in constant time, i.e., it must not depend on the size of the interpreter arguments. This suggests the use of an underlying interpreter that represents values in a compressed way, e.g., using the value-number technique described in \citet[][]{Aho:2007tt}.
In Algorithm 1, \texttt{subproblem} maps interpreter arguments to network nodes.

{\bf Factorization grain.} There are two ways to determine how much information sharing takes place: (1) While the interpreter may cede control at all recursive calls, it does not need to for our algorithm to be valid. There is a continuum between building a fully factored FSPN and building a tree of random choices without factorization. In our experiments, we have found it advantageous to factor at all calls that correspond to function applications. (2) What information the underlying interpreter passes to its recursive calls affects sharing. In Church, where interpreter arguments consist of expressions and environments, restricting environments to {\em relevant} environments is critical for efficient dynamic programming.

{\bf Algorithm.} Our algorithm  maintains a queue of tasks, initialized to a single task for the first interpreter call. Each task is a tuple of a thunk $f$ (a function without arguments), a previous node $\prevnode$, and an edge weight $\prevweight$ (a probability). While the queue is not empty, the algorithm takes the first task in the queue and evaluates the function call $f()$. There are three types of return values (partial results): subcalls, random choices, and terminal values. We process the value according to its type:

\input{algorithm-figure.tex}

{\em Subcalls} are pairs of a continuation $c$ and an interpreter argument $s$. At subcalls, we build a sum node $\curnode$ that will have one child for each return value of the subcall.

If this is a {\em new subcall}, we add a new root node to the network and add to the queue the task of exploring this subcall, starting from the root node.

If this is a {\em known subcall}, we look up what root node it corresponds to and process all return values known for this subcall. For each such value, we add a product and reference node, and add to the queue the task of continuing evaluation using the continuation $c$.

Finally, we store in \texttt{callbacks} for the root node $r$ the continuation $c$ together with $\curnode$ such that, when new return values for the subcall are found, we can continue building the network in the current context.

{\em Random choices} are tuples of a continuation $c$, values $\vec{v}$, and probabilities $\vec{p}$. We add a sum node $\curnode$ and enqueue a call to the continuation for each value in $\vec{v}$.

{\em Terminal values} cause indicator nodes to be built. If a value is new for the current subproblem $r$, we notify all contexts waiting for return values under $r$ by calling \texttt{ProcessTerminal} once for each callback associated with $r$.

\subsection{Solving FSPNs}

A FSPN corresponds to a system of equations (Section \ref{sec:fspn}). For probabilistic programs, this system tends to be sparse, reflecting the fact that, in general, most interpreter calls that occur in the process of enumerating a given program do not depend on most other calls. We therefore cluster the equations into strongly connected components and solve the clusters of equations in topological order. Computing a topological order of strongly connected components is linear in the size of the graph \citep{tarjan:2204443}. By solving in topological order we know that all probabilities required to compute the solution of a component have been computed once we reach this component.

In our examples, we use a simple substitution-based equation simplifier, fixed-point iteration, and Newton's method to solve these equations. Exploring the use of other solution methods is a potential venue for future performance improvements.

\section{EMPIRICAL EVALUATION}

In this section, we describe three situations where we have found generic dynamic programming to be useful: teaching probabilistic models, research in computational cognitive science, and analysis of multi-agent reasoning in game-theoretic situations. We present an example for each of these situations and compare dynamic programming to other inference algorithms.

In {\bf teaching probabilistic models}, we usually aim to present modeling and inference patterns before we discuss the internals of inference algorithms, since the former provide motivation for the latter. The {\em Probabilistic Models of Cognition} tutorial by \citet[][]{ChurchWiki} follows this approach and has been used in graduate classes at MIT and Stanford. Since implementations of probabilistic programming languages supply universal inference algorithms, it is possible to nonetheless allow students to experiment with models and solve exercises. 

However, the performance of existing ``universal'' algorithms strongly depends on the structure of the models they are applied to, even for models with a very small number of variables. Rejection sampling is only feasible as long as we do not condition on low-probability events; MCMC requires that the distribution does not have modes that are isolated with respect to the proposal structure of the algorithm.

Even in cases where sampling is feasible, it poses a challenge to students: it can be difficult to distinguish approximation noise from systematic inference patterns. For Metropolis-Hastings in the space of program traces \citep{Goodman:2008uq, Wingate:2011ul}, quantitative analysis of mixing times does not exist, hence analysis of convergence can be difficult even for experts; students' lack of background knowledge exacerbates this effect.

\input{rope-code-figure.tex}

\begin{figure*}[t]
  \centering
  \includegraphics{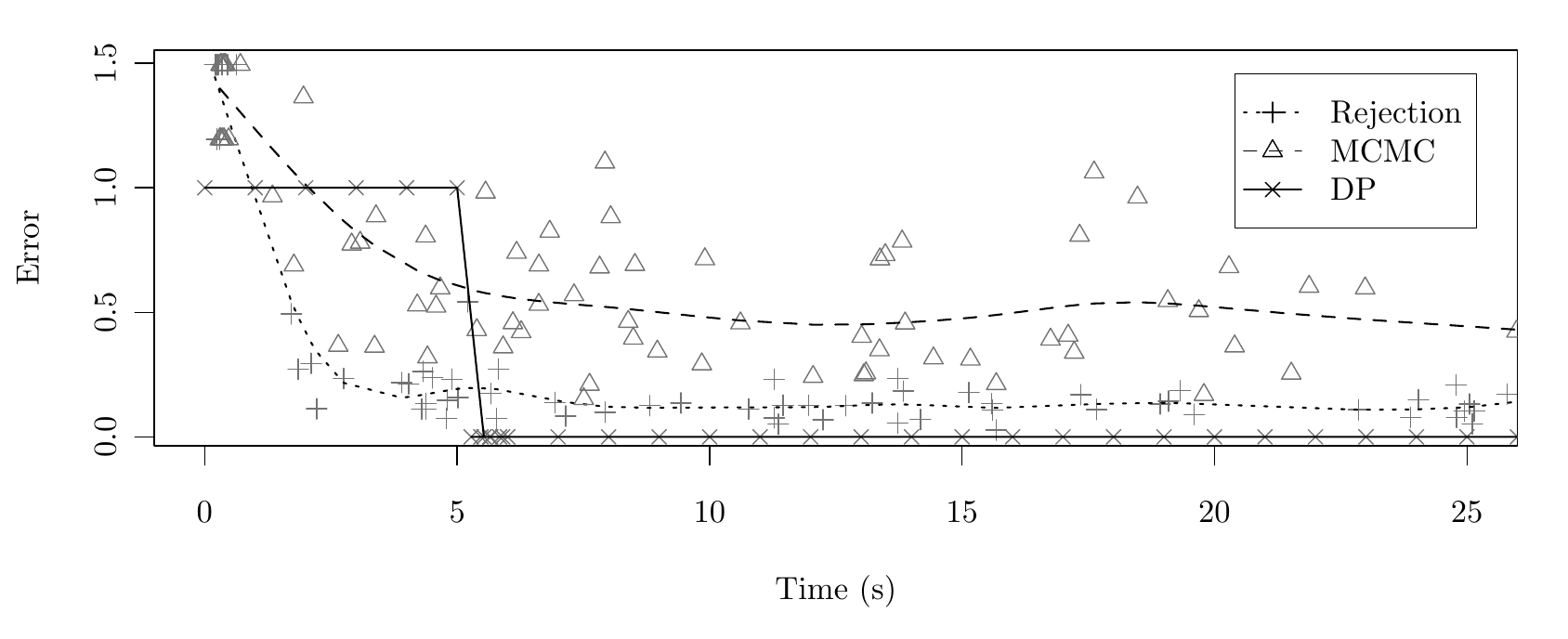}\vskip -1em
  \caption{{\bf Convergence to true distribution for the rope-pulling model.} Each point represents the L1 error between the estimated and true distribution for a given runtime and algorithm. While all algorithms eventually converge to the correct distribution for this model used in teaching, the only algorithm that quickly provides a {\em precise} answer is dynamic programming. In this example, MCMC has difficulty mixing between modes. Rejection computes estimates using very few samples due to a low-probability condition.}
  \label{fig:teaching_example}
\end{figure*}

For example, consider the {\em rope-pulling game} (Figure \ref{fig:rope-model}), a simple probabilistic program without nested conditioning.
Figure \ref{fig:teaching_example} shows how the L1 error between the estimated and true posterior distribution develops over time for rejection, MCMC, and dynamic programming. While MCMC has difficulty mixing between modes, and while rejection computes estimates using very few samples due to a low-probability condition, dynamic programming deterministically returns the exact answer after about $6$ seconds.

In {\bf cognitive science research}, we wish to quickly explore a wide range of model variations. While the model prototypes used in research have a tiny number of variables compared to the state of the art in machine learning, they are structurally complex and use features such as mutual recursion, nested conditioning, and stochastic higher-order functions. Probabilistic programming makes it possible to explore this space without building custom inference algorithms.

\begin{figure*}[t]%
    \begin{subfigure}[b]{0.55\textwidth}%
    \begin{lstlisting}
(define (speaker access state depth)
  (query
   (define sentence (sentence-prior))
   sentence
   (equal? (belief state access)
           (listener access sentence depth))))

(define (listener sp-access sentence depth)
  (query
   (define state (state-prior))
   state
   (if (= 0 depth)
       (sentence state)
       (equal? sentence
               (speaker sp-access state
                        (- depth 1))))))
\end{lstlisting}
\vskip 2em
  \label{fig:probprog_implicature}
  \end{subfigure}%
    \begin{subfigure}[b]{0.35\textwidth}
      \raisebox{-3em}{
      \includegraphics{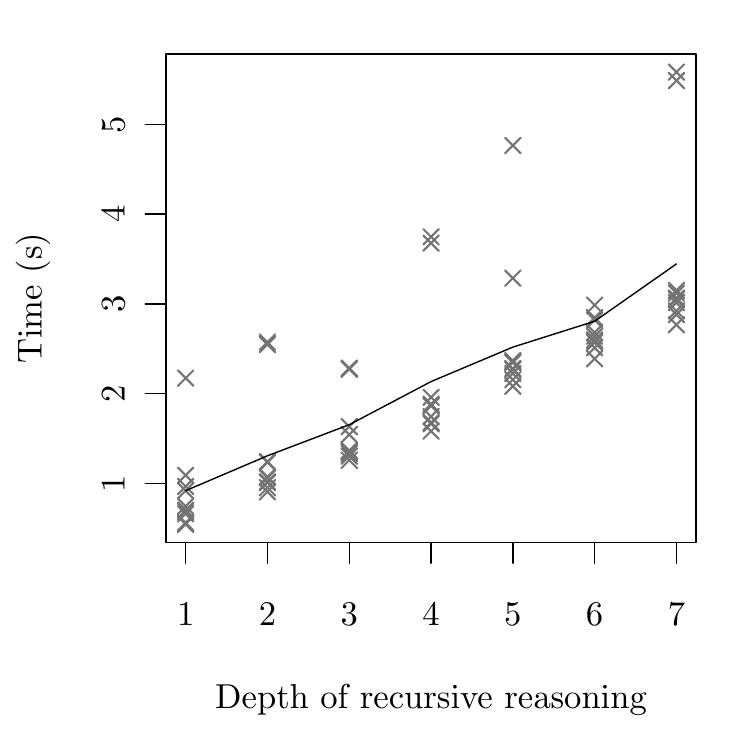}
      }
  \label{fig:implicature_results}
  \end{subfigure}
  \vskip -1em
  \caption{{\bf Increase in dynamic programming inference time as a function of nested conditioning depth.} For this model used in cognitive science research, dynamic programming makes it possible to explore nested recursive conditioning with linear growth of inference time in the depth of recursion. Each point on the plot corresponds to a run of our algorithm on the model with a given depth. For rejection and MCMC over rejection, expected inference time grows exponentially.}
  \label{fig:science_example}
\end{figure*}

The caveat that the performance of current sampling-based algorithms strongly depends on models, even in small state spaces, applies here as well.
For example, \citet{ndg+ast:cogsci2012} proposed a model of language understanding based on the idea that listeners assume that speakers choose their utterances approximately optimally, and that listeners interpret an utterance by using Bayesian inference to ``invert'' this model of the speaker.  Figure \ref{fig:science_example} shows part of a model of this type that predicts an interaction between the speaker's state of knowledge and the listener's interpretation of scalar implicatures (e.g., ``some'' implies ``not all''). Using dynamic programming, the time it takes to compute the marginal distribution for this model grows linearly in the depth of recursive reasoning, whereas for current sampling techniques, inference time grows exponentially.

Moreover, some of the model features that are of research interest do not easily fit into the sampling framework.
For example, in softmax-optimal decision-making, an action $a$ is chosen according to exponentiated expected utility under a belief distribution $P(s)$, i.e.,  $P(a) \propto \exp \left ( \alpha \mathbb{E}_{P(s)}[U(a;s)] \right )$. A direct translation into a probabilistic language with sampling semantics seems to require additional programming constructs that reify distributions. Such constructs can be provided more easily in the setting of exact inference.

The analysis of {\bf multi-agent reasoning in game-theoretic situations} shares many properties with cognitive science research, but places even more emphasis on multiply nested conditioning. This commonly rules out existing sampling-based algorithms. At the same time, enumeration is often not an option either, since exploiting shared structure is critical in reducing the state space to tractable size: in the analysis of multiple agents thinking about one another, we can share computation between all agents, actual and counterfactual, that are modeled as being in the same state of mind.

\begin{figure*}
  \centering
  \includegraphics[scale=.8]{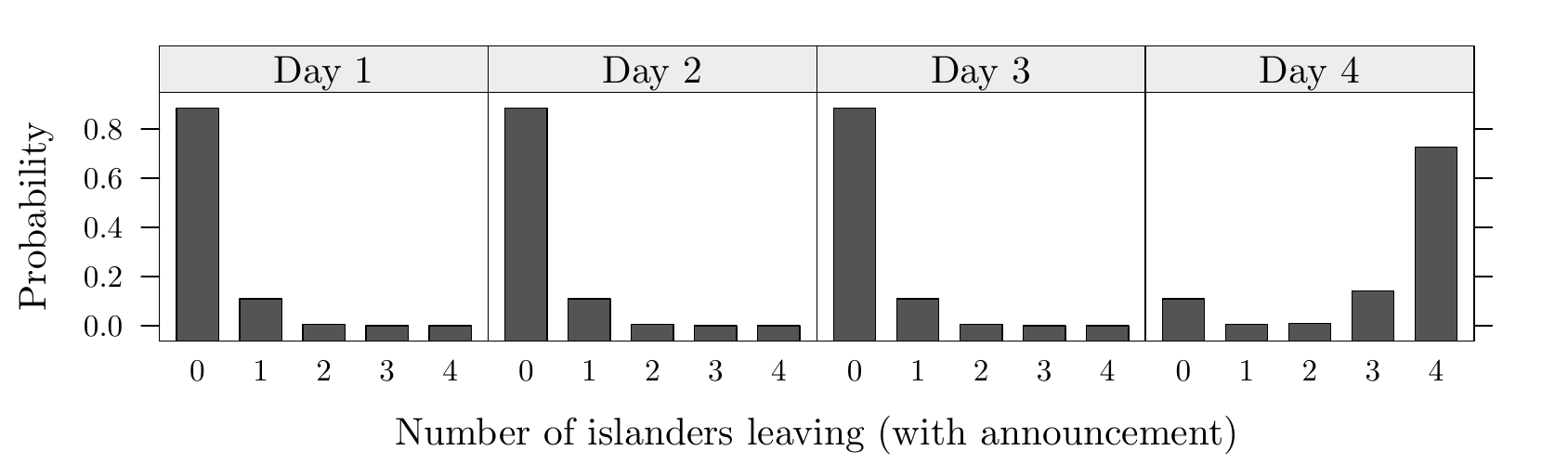}  
  \caption{While the blue-eyed islanders puzzle is challenging for all generic inference algorithms, dynamic programming allows predictions for small population sizes that are already intractable for MCMC and rejection. The figure shows results for population size $4$, all islanders blue-eyed: on the 4th day, it is highly likely that all islanders decide to leave.}
  \label{fig:gametheory_example}
\end{figure*}

As a particularly difficult example, consider the ``blue-eyed islanders'' puzzle, a well-known problem in epistemic logic \citep{Tao2008}. The setup is as follows: There is a tribe on a remote island. Out of the $n$ people in this tribe, $m$ have blue eyes. Their religion forbids them to know their own eye color, or even to discuss the topic. Therefore, everyone sees the eye color of every other islander, but does not know their own eye color. If an islander discovers their color, they have to publicly announce this at noon and leave the island. All islanders are highly logical. One day, a foreigner comes to the island and---speaking to the entire tribe---he says: ``At least one of you has blue eyes.'' What happens next? Results for a stochastic version of this puzzle are shown in Figure \ref{fig:gametheory_example}.

The difficulty of this model stems from the fact that each day, every islander reasons about the reasoning of all of the other islanders on the previous day, and that their reasoning must again include all islanders' reasoning on the day before the previous day, etc. However, due to the symmetry of the setup, all islanders with blue eyes and all islanders without blue eyes do the same computation on any given day. Their computations are merged by our algorithm, which makes exact inference feasible for small populations.

\section{RELATED WORK}

Our algorithm is related to and inspired by a long tradition of algorithms which use dynamic programming to exploit reusable structure in the natural language processing, logic programming, and functional programming literatures. For example, it is known that, in general, exactly solving problems such as marginalization for arbitrary recursive programs leads to systems of nonlinear equations \citep[see, e.g., comments in][]{eisner.j:2005}. \citet{klein.d:2001a} exploit strongly connected components of the computation graph for PCFGs to perform efficient exact marginalization in a way similar to the present algorithm.  It is beyond the scope of this paper to review the many connections with individual algorithms presented in the literature. Instead, we focus on three systems which attempt use dynamic programming to provide general inference algorithms for universal, probabilistic (or, more generally, weighted) programming languages: IBAL, PRISM, and Dyna.

The most closely related system to the present work is the functional programming language IBAL, a probabilistic variant of ML \citep{pfeffer.a:2001}. IBAL provides an exact marginalization algorithm for discrete probabilistic models, which is based on a generalization of variable elimination applied to computation graphs.
The graph used by this algorithm also exploits sharable subcomputations across the evaluation of the probabilistic program. However, the present algorithm is more general than the IBAL algorithm in an important way. The IBAL algorithm relies on {\it acyclic} computation graphs; this is equivalent to the requirement that the computation be {\em evidence-finite} \citep{koller.d:1997}---there must only be a finite number of computations which can give rise to the observed evidence. By contrast, our algorithm handles many cases of {\em evidence-infinite} computation. For example, the simple recursive program shown in Figure \ref{fig:probprog_recursive}a, which has finite support \{{\tt true}, {\tt false}\} but an infinite number of computations which give rise to each support value, cannot be marginalized by IBAL, but is correctly handled by our algorithm. Practical examples of such evidence-infinite computations
include the nested-query models for multi-agent reasoning that we have described above.

Another system which is similar to the present work is PRISM, a probabilistic generalization of Prolog, which also makes use of dynamic programming to provide a general inference algorithm. Although PRISM is able to recover many standard algorithms for problems such as PCFG estimation (e.g., the inside-outside algorithm), like IBAL, it cannot handle evidence-infinite computations \citep{Sato:09aa}.

A somewhat different approach is Dyna \citep{eisner.j:2005}, a programming language for expressing weighted deductive logic programs. Dyna makes use of generalizations of parsing-as-deduction \citep{shieber.s:1995} and semi-ring parsing \citep{goodman.j:1999} to compile weighted logic programs into highly optimized dynamic programs.  Dyna differs from our algorithm in the target level of abstraction. Our algorithm is focused on the problem of rapid prototyping of models for which no standard dynamic programming algorithm exists. The programmer simply provides an interpreter, and our algorithm automatically exploits whatever sharing is exposed by the structure of the recursive calls made in the process of computing the marginal distribution for a particular model. By contrast, Dyna is a language for abstractly expressing {\it specific} dynamic programming \emph{algorithms} and compiling these algorithms to highly efficient code. It allows the programmer lower-level control over algorithm specification, but it also \emph{requires} the programmer to specify these algorithmic details.

\section{CONCLUSION}

We have developed a dynamic programming algorithm for exact inference in probabilistic programs. We have illustrated how this algorithm aids the use of probabilistic programs in teaching and research. Future work includes incorporating techniques from other approaches to dynamic programming (such as evidence propagation from IBAL, and efficient code generation from Dyna) and exploring techniques for approximate dynamic programming.

\subsubsection*{Acknowledgements}

The authors would like to thank Daniel Roy and Timothy O'Donnell for helpful discussions and contributions to early stages of this project. This work was supported by ONR grant N00014-09-0124.

\bibliographystyle{abbrvnat}
\bibliography{dppp}

\end{document}

%% file: algorithm-figure.tex
\begin{figure}[t]
\begin{minipage}{1.0\linewidth}
\hrule height2pt \vskip 0.04in
\textbf{Algorithm 1: Compiling probabilistic programs to factored sum-product networks}
\vskip 0.04in \hrule height1pt \vskip 0.04in

\begin{algorithmic}
  \small
\Procedure{BuildFSPN}{$\Interpreter$, $\initstate$}

  \State $G$ = Graph()
  \State $r$ = $G$.addNode({\em root})
  \State $Q$ = [$(\lambda.\Interpreter(\initstate), r, 1.0)$]
  \State terminals, callbacks, subproblem = \{\}, \{\}, \{\}

  \While{$Q$ {\bf is} not empty}

  \State $(f, \prevnode, \prevweight)$ = $Q$.pop()
  \State $x$ = $f()$

  \If{$x$ {\bf is a} value $v$}
  \State $\curnode$ = $G$.addNode({\em indicator}, $v$)
  \State $r$ = $G$.root[$\prevnode$]
  \If{$v$ $\notin$ terminals[$r$]}
    \ForAll{$(n', c)$ {\bf in} callbacks[$r$]}
      \State \Call{processTerminal}{$G$, $Q$, $r$, $v$, $n'$, $c$}
    \EndFor
    \State terminals[$r$].add($v$)    
  \EndIf

  \ElsIf{$x$ {\bf is a} random choice $(c, \vec{v}, \vec{p})$}
  \State $\curnode$ = $G$.addNode({\em sum})
  \ForAll{$v,p \in \vec{v}, \vec{p}$}
   \State $Q$.enqueue($\lambda.c(v)$, $\curnode$, $p$)
  \EndFor

  \ElsIf{$x$ {\bf is a} subcall $(c, s)$}
  \State $\curnode$ = $G$.addNode({\em sum})
  \If{$s \notin $ subproblem.keys()}
    \State $r$ = $G$.addNode({\em root})
    \State subproblem[$s$] = $r$    
    \State $Q$.enqueue($\lambda.\Interpreter(s)$, $r$, 1.0)
  \Else
    \State $r$ = subproblem[$s$]
    \ForAll{$v \in$ terminals[$r$]}
      \State \Call{processTerminal}{$G$, $Q$, $r$, $v$, $\curnode$, $c$}
    \EndFor
  \EndIf
  \State callbacks[$r$].add($(\curnode, c)$)      
  \EndIf
  \State $G$.addEdge($\prevnode$, $\curnode$, $\prevweight$)
  \EndWhile
\Return G
\EndProcedure
\State
\Procedure{ProcessTerminal}{$G$, $Q$, $\rootnode$, $v$, $\prevnode$, $c$}
  \State $\prodnode$ = $G$.addNode({\em product})
  \State $\refnode$ = $G$.addNode({\em ref}, $\rootnode$, $v$)
  \State $G$.addEdge($\prevnode$, $\prodnode$, 1.0)
  \State $G$.addEdge($\prodnode$, $\refnode$, 1.0)
  \State $Q$.enqueue($\lambda.c(v)$, $\prodnode$, 1.0)
\EndProcedure
\end{algorithmic}%
\hrule height1pt \vskip 0.04in
\vskip -1em
\end{minipage}
\end{figure}%

%% file: rope-code-figure.tex
\begin{figure}[t!]
\begin{lstlisting}[mathescape=true]
(query

  ;; Generative model
  (define team1 (list 0 1))
  (define team2 (list 2 3))
  (define strengths
    (repeat 4 ($\lambda$ () (if (flip) 10 5))))
  (define (strength person)
    (list-ref strengths person))
  (define (lazy person)
    (flip (/ 1 3)))
  (define (total-pulling team)
    (sum
     (map ($\lambda$ (person)
            (if (lazy person)
                (/ (strength person) 2)
                (strength person)))
          team)))
  (define (winner team1 team2)
    (if (< (total-pulling team1)
           (total-pulling team2))
           'team2
           'team1))

  ;; Query expression
  (list (strength 0) (strength 1))

  ;; Condition
  (and
    (eq? 'team1 (winner team1 team2))
    $\dots$
    (eq? 'team2 (winner team1 team2))))
\end{lstlisting}
\caption{The rope-pulling game, a simple generative model used in teaching probabilistic modeling.}
\label{fig:rope-model}
\end{figure}